\documentclass[12pt]{article}

\usepackage{amsmath,amssymb,amsthm}
\usepackage{graphicx}
\usepackage[hidelinks]{hyperref}
\usepackage{geometry}
\usepackage{bm}
\usepackage[utf8x]{inputenc}
\usepackage{amssymb,epstopdf,verbatim,gensymb,amsmath,amsthm}
\usepackage{nccmath, mathtools,bm,url}
\usepackage{textcomp}
\usepackage{amsmath}
\usepackage{float,epsfig}
\usepackage{subfig,boldline}
\usepackage{amsfonts,color}
\usepackage{tabularx,longtable,adjustbox,siunitx,float,multirow,graphicx,booktabs}

\usepackage{doi}
\graphicspath{{./figs/}}

\title{\textbf{Untrained CNN and coefficient-space regularization for joint reconstruction in photoacoustic tomography}}

\author{Alfio Borz\`i \thanks{alfio.borzi@uni-wuerzburg.de, Institut f\"ur Mathematik, Universit\"at W\"urzburg, Emil-Fischer-Strasse 30, 97074 W\"urzburg, Germany.} \and
Anwesa Dey \thanks{anwesary@math.utah.edu, Department of Mathematics, The University of Utah, John A. Widtsoe Building 
155 S 1400 E, Salt Lake City, 84112, Utah, USA. } 
\and Souvik Roy \thanks{souvik.roy@uta.edu, Department of Mathematics, The University of Texas at Arlington, 411 S. Nedderman Drive, Arlington, 76019, Texas, USA. }}

\date{\today}

\begin{document}

\maketitle

\begin{abstract}

The joint reconstruction of the initial pressure and spatially varying
sound speed in photoacoustic tomography from boundary measurements is
considered. A regularized data-driven framework is proposed in which the
unknown fields are represented in finite-dimensional wavelet spaces and
fixed nonlinear features generated by an untrained convolutional neural
network (CNN) are used to reconstruct the initial-pressure coefficients
through Tikhonov-regularized least squares. A learned coefficient-space
prior is subsequently used to infer the squared sound-speed coefficients.
Numerical experiments demonstrate accurate reconstruction on held-out
data and strong generalization to previously unseen phantom geometries.
The frozen reconstruction map also remains stable under measurement
noise not represented during training, with only minor changes in
reconstruction accuracy for noise levels up to $10\%$. In comparison
with the supervised trained-CNN approach considered in this work, the
untrained representation yields substantially smaller reconstruction
errors while requiring approximately one-sixth of the computational
time. These results demonstrate that fixed random convolutional features
combined with regularized coefficient-space learning provide an
efficient and robust approach to joint reconstruction in photoacoustic tomography.

\end{abstract}



\section{Introduction}

Photoacoustic tomography (PAT) is a hybrid imaging modality that combines optical contrast 
with ultrasonic spatial resolution \cite{xu2006photoacoustic,beard2011biomedical}. 
Following the absorption of a short optical pulse, thermoelastic expansion 
generates an initial acoustic pressure distribution inside the tissue. 
The resulting pressure wave propagates through the acoustic medium and is recorded 
by detectors located on or near the boundary of the imaging domain. 
The reconstruction of the initial pressure from these measurements constitutes the acoustic
inverse problem in PAT and has been studied using analytical, time-reversal, 
model-based, and iterative reconstruction techniques
\cite{cox2009photoacoustic,beard2011biomedical,arridge2016adjoint,dey2024high}.

A central difficulty in photoacoustic reconstruction is that acoustic propagation 
depends on the spatially varying properties of the medium. In particular, the 
sound speed is generally heterogeneous in biological
tissue. Assuming an inaccurate or spatially homogeneous sound speed can
produce errors in acoustic travel times and consequently lead to
mislocalization, blurring, and distortion of the reconstructed initial
pressure. Although many conventional PAT reconstruction methods assume
that the sound speed is known, simultaneous estimation of the initial
pressure and sound speed has therefore received considerable attention
\cite{stefanov2009thermoacoustic,huang2013full,matthews2018parameterized,
matthews2017joint}. The resulting problem is substantially more
difficult than reconstruction of the initial pressure alone because both
unknown fields influence the same measured acoustic wave field.

From the inverse-problem perspective, joint reconstruction is also
fundamentally more delicate than the conventional PAT problem. 
Stefanov and Uhlmann \cite{stefanov2009thermoacoustic,
stefanov2013instability} established important analytical results for
thermoacoustic tomography with variable sound speed and, in the
simultaneous source-and-speed setting, showed that the corresponding
linearized inverse problem is unstable in Sobolev scales. Numerical
investigations have similarly demonstrated the strong coupling between
the two unknown fields \cite{huang2013full}. Consequently, prior
information, regularization, or reduced-dimensional parameterizations
are needed to obtain useful joint reconstructions. For example,
parameterized representations of the sound-speed distribution have been
employed to stabilize the joint problem
\cite{matthews2018parameterized}, while additional ultrasound
measurements have been incorporated to provide complementary information
for joint recovery of the initial pressure and sound speed
\cite{matthews2017joint}.

Data-driven reconstruction has emerged as an alternative and
complementary approach to conventional model-based inversion. Neural
networks have been investigated for direct reconstruction, artifact
removal, limited-view imaging, undersampling correction, and
model-enhanced reconstruction in PAT
\cite{hauptmann2020deep,grohl2021deep,dispirito2021sounding,roy2025pinn}.
Such approaches can provide rapid inference after training and can
incorporate statistical information that is difficult to express through
traditional regularization functionals. At the same time, fully trained
networks may require large training sets and substantial computational
effort, and their performance can depend strongly on the correspondence
between the training distribution and the objects encountered during
inference \cite{hauptmann2020deep,grohl2021deep}. This motivates the
consideration of fixed random nonlinear representations, in which the
convolutional feature map itself does not require optimization and only
the subsequent reconstruction map is learned from the data. Such an
approach retains nonlinear feature extraction while shifting the
trainable component to a regularized finite-dimensional regression
problem. These considerations motivate reconstruction strategies in
which data-driven representations are combined with explicit
regularization and low-dimensional statistical prior information.

Other approaches have exploited structural feature coupling for joint
initial-pressure and sound-speed reconstruction, as well as trained
deep networks embedded in model-based iterations
\cite{cai2019feature,shan2019simultaneous}. Untrained neural networks
have also been investigated in PAT, although in the distinct setting of
compressed-sensing reconstruction based on deep image priors
\cite{lan2021compressed}. More recently, simultaneous reconstruction from multiple
initial-pressure data sets sharing the same sound-speed distribution has
been investigated as another mechanism for reducing the ill-posedness
of the joint problem \cite{suhonen2025multiple}.

In the present work, we consider the joint inverse problem of recovering
the initial pressure $f$ and the spatially varying sound speed $c$ from
boundary measurements of the acoustic field. Our approach is motivated
by the numerical observation that, for the class of problems considered
here, the inverse map from the boundary measurements to the initial
pressure is substantially easier to approximate than the corresponding
map to the sound speed. Rather than reconstructing the two fields
independently, we therefore adopt a sequential strategy in which the
initial pressure is reconstructed first and its coefficient vector is
subsequently used to infer the statistically predictable component of
the squared sound speed.

The proposed framework combines several complementary forms of
regularization. First, both unknown fields are represented in
finite-dimensional wavelet spaces, thereby restricting the inverse
problem to multiscale coefficient spaces. Second, the measured
Dirichlet and Neumann boundary traces are mapped to nonlinear features
using a convolutional neural network whose filters are randomly
generated and then held fixed. The pressure coefficients are recovered
from these features through Tikhonov-regularized least squares, avoiding
the need to train the convolutional representation. Third, a regularized
coefficient-space prior learned from the training ensemble maps the
reconstructed pressure coefficients to the squared sound-speed
coefficients. The statistical model used to construct the ensemble
includes an independent sound-speed component, so this prior does not
impose a deterministic relation between the two unknown fields. The
distinctive feature of our approach is therefore not any single
component in isolation, but their combination within one regularized
reconstruction architecture. To our knowledge, joint PAT reconstruction
has not previously been formulated by combining fixed random
convolutional measurement features, Tikhonov-regularized recovery in
wavelet coefficient space, and a learned statistical prior between the
pressure and squared-sound-speed coefficients.

The numerical study addresses four principal questions: whether the
coefficient-space framework can accurately reconstruct both unknown
fields on held-out data, whether the reconstruction generalizes to
phantom geometries not represented in the training ensemble, whether
the frozen reconstruction map remains stable when the boundary
measurements are contaminated by noise not represented during training,
and whether training the convolutional feature extractor provides an
advantage over the proposed fixed random representation. For the
supervised trained-CNN comparator considered in this work, the fixed
random features yield substantially smaller reconstruction errors for
both the initial pressure and sound speed while requiring approximately
one-sixth of the total computational time. Accurate reconstructions on
previously unseen geometries further indicate that the method is not
restricted to reproducing the particular structures represented in the
training set. Moreover, the reconstruction accuracy remains essentially
unchanged for measurement-noise levels up to $10\%$, including for the
previously unseen phantom geometries.

The remainder of the paper is organized as follows.
Section~\ref{sec:inverse_problem} formulates the forward model and the
joint inverse problem. Section~\ref{sec:wavelet} introduces the
finite-dimensional wavelet representations, followed by the statistical
model for the coupled acoustic parameters and the coefficient-space
prior. Section~\ref{sec:cnn} develops the fixed convolutional feature
map and Tikhonov-regularized reconstruction of the initial pressure.
In Section~\ref{sec:numerics}, we present the results and discussion of
numerical experiments evaluating reconstruction accuracy, robustness to
measurement noise, generalization to unseen geometries, and comparison
with a supervised trained CNN. We end with a section on conclusions.

\section{Formulation of the joint inverse problem}
\label{sec:inverse_problem}

Let $\Omega\subset\mathbb{R}^d$ be a bounded imaging domain and let
$c:\Omega\rightarrow\mathbb{R}_+$ denote the spatially varying acoustic
sound speed. The propagation of the photoacoustic pressure $p(x,t)$ is
modeled by
\begin{equation}
    p_{tt}(x,t)=c^2(x)\Delta p(x,t),
    \qquad (x,t)\in\Omega\times(0,T),
    \label{eq:wave}
\end{equation}
subject to the initial conditions
\begin{equation}
    p(x,0)=f(x),\qquad
    p_t(x,0)=0,
    \label{eq:initial}
\end{equation}
where $f$ is the unknown initial pressure distribution.

We assume that measurements of the acoustic field are available on the
boundary of the imaging domain. Specifically, the measured data are the
Dirichlet and Neumann traces
\begin{equation}
    g(x,t)=p(x,t),\qquad
    h(x,t)=\frac{\partial p}{\partial\nu}(x,t),
    \qquad (x,t)\in\partial\Omega\times(0,T),
    \label{eq:measurements}
\end{equation}
where $\nu$ denotes the outward unit normal to $\partial\Omega$. 
For convenience, we introduce the squared sound speed
\begin{equation}
    q(x)=c^2(x).
\end{equation}
Let $\mathcal{F}$ denote the forward observation operator associated
with \eqref{eq:wave}--\eqref{eq:measurements}. Thus,
\begin{equation}
    \mathcal{F}(f,q)=(g,h).
    \label{eq:forward_operator}
\end{equation}
The joint inverse problem considered in this work is therefore to
determine the pair $(f,q)$ from the measured boundary data,
\begin{equation}
    (g,h)\longmapsto(f,q).
    \label{eq:inverse_map}
\end{equation}
In contrast to the conventional photoacoustic inverse problem in which
the sound speed is assumed to be known, both arguments of the forward
operator are unknown in \eqref{eq:forward_operator}. The initial
pressure determines the initial state of the acoustic field, whereas
$q$ controls its subsequent propagation. Consequently, perturbations
in either quantity modify the measured boundary traces, producing a
strongly coupled nonlinear inverse problem.

Our objective is not to invert $\mathcal{F}$ directly in the full
physical-space discretization. Instead, we construct a regularized
finite-dimensional approximation of the inverse map. The unknown fields
are first represented in wavelet spaces, and the boundary measurements
are mapped to a fixed nonlinear feature space. Tikhonov-regularized
regression is then used to recover the pressure coefficients, while a
coefficient-space prior transfers the statistically predictable
information in the reconstructed pressure to the squared sound speed.
The required finite-dimensional representation of the unknown fields is
introduced next.

\section{Wavelet representation of the unknown fields}
\label{sec:wavelet}

We represent the unknown initial pressure and squared sound speed in
finite-dimensional multiscale wavelet spaces. Besides reducing the
dimension of the reconstruction problem, this parameterization acts as
a representation regularization by restricting the admissible
reconstructions to prescribed finite-dimensional spaces. Wavelets are
particularly suitable for this purpose because their spatial
localization and multiscale structure allow both smooth regions and
localized interfaces to be represented efficiently \cite{daubechies1992ten,mallat2009wavelet}.

Let
\[
V_f=\operatorname{span}\{\psi_1,\ldots,\psi_{N_f}\}
\]
denote the finite-dimensional space used for the initial pressure. We
seek an approximation of the form
\begin{equation}
    f(x)=\sum_{i=1}^{N_f}\beta_i\psi_i(x),
    \label{eq:f_wavelet}
\end{equation}
where
\[
    \beta=(\beta_1,\ldots,\beta_{N_f})^T\in\mathbb{R}^{N_f}
\]
is the coefficient vector to be reconstructed.

For the acoustic parameter, we work with the squared sound speed
\begin{equation}
    q(x)=c(x)^2,
    \label{eq:q_def}
\end{equation}
which appears directly as the spatially varying coefficient in the
wave equation. Let
\[
V_q=\operatorname{span}\{\phi_1,\ldots,\phi_{N_q}\}.
\]
We represent
\begin{equation}
    q(x)=\sum_{j=1}^{N_q}\gamma_j\phi_j(x),
    \label{eq:q_wavelet}
\end{equation}
with
\[
    \gamma=(\gamma_1,\ldots,\gamma_{N_q})^T
    \in\mathbb{R}^{N_q}.
\]

The joint inverse problem is therefore replaced by the
finite-dimensional recovery of the coefficient pair
\[
    (\beta,\gamma)\in
    \mathbb{R}^{N_f}\times\mathbb{R}^{N_q}.
\]
If $\mathcal{S}_f$ and $\mathcal{S}_q$ denote the corresponding wavelet
synthesis operators, then
\[
    f=\mathcal{S}_f(\beta),
    \qquad
    q=\mathcal{S}_q(\gamma),
    \qquad
    c=\sqrt{\mathcal{S}_q(\gamma)}.
\]
Thus, the reconstruction procedure developed below can be viewed as
constructing a regularized map from the measurement data $(g,h)$ to the
coefficient pair $(\beta,\gamma)$ rather than directly to the
high-dimensional physical fields.

In the numerical experiments, tensor-product Haar wavelets are used.
Their compact support and multiscale structure provide a simple
representation of the piecewise regular structures and sharp
transitions appearing in the generated phantoms. Different wavelet
families may be incorporated without changing the reconstruction
framework. In particular, smoother bases such as Daubechies or
biorthogonal wavelets may be advantageous when additional regularity of
the reconstructed fields is expected.

\section{A coupled generative model for the acoustic parameters}
\label{sec:generative}

To construct the training ensemble, we introduce a statistical coupling
between the initial pressure and the squared sound speed. This coupling
is a modeling assumption used to generate the data and to construct the
coefficient-space prior; it is not imposed as a deterministic relation
between the two unknown fields.

We assume that part of the spatial variability of the initial pressure
and sound speed arises from common underlying structure. At the same
time, the sound speed is allowed to contain an independent component
that cannot be inferred from the pressure field alone.

Let
\[
\xi=(\xi_1,\ldots,\xi_{N_f})^T
\]
be a random generating coefficient vector. An initial pressure
realization is generated from
\[
f_{\rm raw}(x)
=
\sum_{i=1}^{N_f}\xi_i\psi_i(x).
\]
Since the photoacoustic initial pressure is nonnegative, we normalize
the field according to
\begin{equation}
f_{01}(x)
=
\frac{f_{\rm raw}(x)-\min f_{\rm raw}}
{\max f_{\rm raw}-\min f_{\rm raw}},
\label{eq:f01}
\end{equation}
so that $0\leq f_{01}(x)\leq1$. The physical initial pressure is then
defined by
\begin{equation}
f(x)=A_f f_{01}(x),
\label{eq:fphysical}
\end{equation}
where $A_f>0$ denotes the pressure amplitude. 

To introduce variability in the sound speed that is independent of
$f$, a second random coefficient vector
\[
\alpha=(\alpha_1,\ldots,\alpha_{N_q})^T
\]
is generated independently. It defines the auxiliary field
\[
\varepsilon_{\rm raw}(x)
=
\sum_{j=1}^{N_q}\alpha_j\phi_j(x),
\]
which is normalized as
\begin{equation}
\varepsilon(x)
=
\frac{\varepsilon_{\rm raw}(x)}
{\|\varepsilon_{\rm raw}\|_\infty},
\label{eq:epsilon}
\end{equation}
so that $|\varepsilon(x)|\leq1$.

The squared sound speed is generated according to
\begin{equation}
q(x)
=
c_0^2
+
K\bigl(f_{01}(x)-\overline{f}_{01}\bigr)
+
\sigma_\varepsilon\varepsilon(x),
\label{eq:qgeneration}
\end{equation}
where
\[
\overline{f}_{01}
=
\frac{1}{|\Omega|}
\int_\Omega f_{01}(x)\,dx.
\]
Here, $c_0$ is the reference sound speed, $K$ controls the strength of
the statistical coupling between the two fields, and
$\sigma_\varepsilon$ determines the magnitude of the independent
sound-speed fluctuations. The centering term
$f_{01}-\overline{f}_{01}$ keeps the spatial mean of $q$ close to
$c_0^2$.

To ensure physical admissibility, the generated squared sound speed is
projected onto the interval $[q_{\min},q_{\max}]$:
\begin{equation}
q(x)
\leftarrow
\min\left(q_{\max},
\max\left(q_{\min},q(x)\right)\right),
\label{eq:qprojection}
\end{equation}
where
\[
q_{\min}=c_{\min}^2,
\qquad
q_{\max}=c_{\max}^2.
\]

Before the projection in \eqref{eq:qprojection}, the model
\eqref{eq:qgeneration} can be written as
\begin{equation}
q=q_{\rm corr}+q_{\rm ind},
\label{eq:qdecomposition}
\end{equation}
where
\[
q_{\rm corr}
=
c_0^2+
K\bigl(f_{01}-\overline{f}_{01}\bigr),
\qquad
q_{\rm ind}
=
\sigma_\varepsilon\varepsilon.
\]
The first component is statistically correlated with the initial
pressure, whereas the second is generated independently.

This decomposition is important for the inverse problem. Even exact
knowledge of $f$ does not determine $q$ uniquely within the generative
model because of $q_{\rm ind}$. The objective of the coefficient-space
prior introduced below is therefore not to impose a deterministic
relation $q=q(f)$, but to recover the component of the squared sound
speed that is statistically predictable from the pressure field. The
independent component represents variability that cannot, in general,
be recovered from $f$ alone.

\section{Coefficient-space prior for the sound speed}
\label{sec:coefficient_prior}

The generative model in the previous section introduces statistical
dependence between the initial pressure $f$ and the squared sound speed
$q=c^2$. We exploit this dependence directly in coefficient space.
Rather than reconstructing the two unknown coefficient vectors
independently from the boundary data, we first recover the pressure
coefficients and then use their statistically predictable relation with
the sound-speed coefficients as a prior.

Let
\[
\beta^{(m)}\in\mathbb{R}^{N_f},
\qquad
\gamma^{(m)}\in\mathbb{R}^{N_q},
\qquad m=1,\ldots,M_{\rm tr},
\]
denote the pressure and squared-sound-speed coefficient vectors in the
training set. Define the matrices
\[
B=
\begin{bmatrix}
(\beta^{(1)})^T\\
\vdots\\
(\beta^{(M_{\rm tr})})^T
\end{bmatrix}
\in\mathbb{R}^{M_{\rm tr}\times N_f},
\qquad
\Gamma=
\begin{bmatrix}
(\gamma^{(1)})^T\\
\vdots\\
(\gamma^{(M_{\rm tr})})^T
\end{bmatrix}
\in\mathbb{R}^{M_{\rm tr}\times N_q}.
\]
To allow for an affine relation, we augment the pressure coefficient
matrix by a constant column,
\[
\widetilde B=
\begin{bmatrix}
\mathbf{1} & B
\end{bmatrix}.
\]
We seek a linear coefficient-space map
\[
\gamma^T\approx [1,\beta^T] \, T,
\]
where
\[
T\in\mathbb{R}^{(N_f+1)\times N_q}.
\]
The matrix $T$ is determined from the regularized least-squares problem
\begin{equation}
T
=
\arg\min_{\widetilde T}
\left\{
\|\widetilde B\widetilde T-\Gamma\|_F^2
+
\lambda_T\|L\widetilde T\|_F^2
\right\},
\label{eq:T_prior_min}
\end{equation}
where $\lambda_T>0$ is the regularization parameter and
\[
L=
\operatorname{diag}(0,1,\ldots,1)
\]
ensures that the affine offset is not penalized. 

The corresponding normal equations are
\begin{equation}
\left(
\widetilde B^T\widetilde B+\lambda_T L^T L \right)T
=
\widetilde B^T \Gamma.
\label{eq:T_prior_normal}
\end{equation}
Since $L$ is symmetric and idempotent, $L^T L=L$, and therefore
\begin{equation}
T = \left( \widetilde B^T\widetilde B+\lambda_T L \right)^{-1}
\widetilde B^T\Gamma.
\label{eq:T_prior}
\end{equation}
Thus, the construction of the coefficient-space prior is itself a
Tikhonov-regularized inverse problem.

Once an approximation $\widehat\beta$ of the pressure coefficients has
been obtained from the boundary measurements, the prior prediction of
the squared-sound-speed coefficients is
\begin{equation}
\widehat\gamma^{\,T}
=
[1,\widehat\beta^T]T.
\label{eq:gamma_prior}
\end{equation}
The corresponding physical fields are reconstructed through the wavelet
synthesis operators,
\begin{equation}
f_0=\mathcal{S}_f(\widehat\beta),
\qquad
q_0=\mathcal{S}_q(\widehat\gamma).
\label{eq:prior_fields}
\end{equation}
Physical admissibility is enforced by projection,
\begin{equation}
q_0(x)
\leftarrow
\mathcal{P}_{[q_{\min},q_{\max}]}\bigl(q_0(x)\bigr),
\label{eq:q_prior_projection}
\end{equation}
and the initial sound-speed reconstruction is
\begin{equation}
c_0^{\rm rec}(x)=\sqrt{q_0(x)}.
\label{eq:c_prior}
\end{equation}

The role of \eqref{eq:gamma_prior} is regularizing rather than
deterministic. In particular, the generative model contains an
independent component of $q$ that is not determined by $f$. Hence the
map $T$ does not imply that the sound speed is uniquely determined by
the initial pressure. Instead, it provides a finite-dimensional
statistical prior that extracts the component of $\gamma$ that is
predictable from $\beta$. The component of the squared sound speed that is not statistically predictable from the pressure coefficients therefore represents an
intrinsic limitation of the coefficient-space reconstruction considered
here. Accordingly, the proposed method should be interpreted as a
regularized reconstruction on the statistical model class represented
by the training ensemble, rather than as a solution of the unrestricted
simultaneous inverse problem for arbitrary pairs $(f,q)$.

This construction also reduces the difficulty of the joint inverse
problem. Instead of attempting to recover the high-dimensional pair
$(\beta,\gamma)$ independently from $(g,h)$, the method first estimates
$\beta$ and then restricts the initial estimate of $\gamma$ through the
regularized coefficient relation \eqref{eq:gamma_prior}. The complete
measurement-to-coefficient reconstruction of $\beta$ is developed in
the next section.

\section{Convolutional reconstruction of the initial pressure}
\label{sec:cnn}

The previous sections established that the principal quantity to be
reconstructed directly from the acoustic measurements is the coefficient
vector $\beta$ associated with the initial pressure. Once an approximation
$\widehat{\beta}$ has been obtained, the coefficient-space prior provides
an estimate of the squared sound speed through
\[
\widehat{\gamma}^{\,T} = [1,\widehat{\beta}^{\,T}]T.
\]
The joint inverse problem is therefore approached sequentially: the
pressure coefficients are first reconstructed from the measured acoustic
data, after which the coefficient-space prior provides an initial
reconstruction of the squared sound speed.

For each realization of the acoustic parameters, the wave equation
\eqref{eq:wave} is solved numerically and boundary measurements are
recorded. The resulting data consist of two matrices
\[
g\in\mathbb{R}^{N_b\times N_t},
\qquad
h\in\mathbb{R}^{N_b\times N_t},
\]
representing the Dirichlet and Neumann traces of the acoustic field
along the boundary. The dimensions $N_b$ and $N_t$ denote the number
of boundary measurement locations and temporal sampling points,
respectively. These boundary-time arrays contain the information from
which the finite-dimensional inverse map to the pressure coefficients
is constructed.

Direct regression from the full boundary-time measurements to the
high-dimensional coefficient vector $\beta$ is computationally
expensive. We therefore introduce a fixed nonlinear dimensional
reduction using an untrained convolutional neural network. The use of
random feature maps is motivated by the observation that useful
nonlinear representations can be obtained from randomly generated
features without explicit optimization of the feature map
\cite{saxe2011random,rahimi2008random}.

Let $K_g^{(1)},\ldots,K_g^{(n_f)}$ and
$K_h^{(1)},\ldots,K_h^{(n_f)}$ denote randomly generated convolution
kernels. The kernels and their corresponding biases are generated once
and subsequently held fixed throughout the reconstruction procedure.
In particular, no backpropagation or optimization of the convolutional
filters is performed. The activated feature maps are
\[
A_g^{(k)}
=
\sigma\left(
K_g^{(k)}*g+b_g^{(k)}
\right),
\qquad
A_h^{(k)}
=
\sigma\left(
K_h^{(k)}*h+b_h^{(k)}
\right),
\]
where $*$ denotes convolution and $\sigma\equiv\tanh$ is the nonlinear
activation function. The resulting feature maps are subsequently
downsampled through average pooling.

The pooled feature maps are vectorized and concatenated to form a
feature vector
\[
z\in\mathbb{R}^{N_z}.
\]
Thus, the fixed convolutional feature extractor defines the nonlinear
measurement map
\begin{equation}
(g,h)\longmapsto z.
\label{eq:measurement_feature_map}
\end{equation}
The convolutional network is therefore used here as a fixed nonlinear
feature transformation rather than as a conventionally trained neural
network.

\medskip

To account for a constant offset, we introduce the augmented feature
vector
\[
\widetilde z=
\begin{bmatrix}
1\\ z
\end{bmatrix}
\in\mathbb{R}^{N_z+1}.
\]
Let $\widetilde z^{(m)}$ denote the augmented feature vector associated
with the $m$-th training sample, and let
\[
\beta^{(m)}\in\mathbb{R}^{N_f}
\]
denote the corresponding vector of pressure coefficients. We collect
the training feature vectors into
\[
Z
=
\begin{bmatrix}
(\widetilde z^{(1)})^T\\
(\widetilde z^{(2)})^T\\
\vdots\\
(\widetilde z^{(M)})^T
\end{bmatrix}
\in
\mathbb{R}^{M\times(N_z+1)},
\]
and the corresponding pressure coefficients into
\[
B
=
\begin{bmatrix}
(\beta^{(1)})^T\\
(\beta^{(2)})^T\\
\vdots\\
(\beta^{(M)})^T
\end{bmatrix}
\in
\mathbb{R}^{M\times N_f}.
\]
The reconstruction matrix
\[
W_\beta
\in
\mathbb{R}^{(N_z+1)\times N_f}
\]
is determined through the Tikhonov-regularized least-squares problem
\begin{equation}
W_\beta
=
\arg\min_W
\left\{
\|ZW-B\|_F^2
+
\lambda_\beta\|W\|_F^2
\right\},
\label{eq:ridgebeta}
\end{equation}
where $\lambda_\beta>0$ is the regularization parameter. The
regularization term limits the sensitivity of the coefficient map to
the extracted measurement features and stabilizes the recovery of the
high-dimensional pressure coefficient vector.

The corresponding normal equations are
\begin{equation}
(Z^TZ+\lambda_\beta I)W_\beta
=
Z^TB,
\label{eq:ridgebetanormal}
\end{equation}
and hence
\begin{equation}
W_\beta
=
(Z^TZ+\lambda_\beta I)^{-1}Z^TB.
\label{eq:ridgebetasolution}
\end{equation}
In the numerical experiments reported below, we use
$\lambda_\beta=10^2$.

Once $W_\beta$ has been determined, a new measurement pair $(g,h)$ is 
processed by the same fixed convolutional feature extractor to obtain $z$. 
The pressure coefficients are reconstructed according to 
\begin{equation} 
\widehat{\beta}^{\,T} = \widetilde z^{\,T}W_\beta.
\label{eq:betahat} 
\end{equation}
The corresponding initial-pressure reconstruction is given by 
\begin{equation}
f_0(x)
=
\sum_{j=1}^{N_f}
\widehat{\beta}_j\psi_j(x).
\label{eq:f0}
\end{equation}

The coefficient-space prior is subsequently used to infer the wavelet coefficients of the squared sound speed: 
\begin{equation} 
\widehat{\gamma}^{\,T} = [1,\widehat{\beta}^{\,T}] \, T. 
\label{eq:qreconstruction} 
\end{equation}
The baseline squared sound speed is reconstructed as
\begin{equation}
q_0(x)
=
\sum_{j=1}^{N_q}
\widehat{\gamma}_j\phi_j(x),
\label{eq:q0}
\end{equation}
followed by projection onto the physically admissible interval,
\begin{equation}
q_0
\leftarrow
\mathcal{P}_{[q_{\min},q_{\max}]}(q_0),
\qquad
c_0^{\rm rec}(x)=\sqrt{q_0(x)}.
\label{eq:c0rec}
\end{equation}
The pair $(f_0,c_0^{\rm rec})$ will be referred to as the
coefficient-space baseline reconstruction. Thus, the baseline reconstruction may be summarized as
\[
(g,h)
\longmapsto
z
\longmapsto
\widehat{\beta}
\longmapsto
f_0,
\qquad
\widehat{\beta}
\longmapsto
\widehat{\gamma}
\longmapsto
q_0
\longmapsto
c_0^{\rm rec}.
\]
The construction combines a fixed nonlinear representation of the
boundary measurements with Tikhonov regularization in coefficient
space and the statistical prior relating the two unknown acoustic
fields.

\section{Numerical experiments}
\label{sec:numerics}

We investigate the reconstruction framework developed in
the previous sections using synthetic photoacoustic data. Two sets of numerical
experiments are considered. First, the method is evaluated on training and
held-out test realizations generated from the same statistical phantom model.
This experiment measures reconstruction accuracy within the distribution
used to construct the learned maps. Second, the trained reconstruction
framework is applied to deterministic phantom geometries that are not
represented in the training data. The latter experiment is intended to
assess generalization to previously unseen spatial structures.

\subsection{Training and test experiments}
\label{subsec:training_test}

The computational domain is chosen as $\Omega=[-1,1]^2,$
and a uniform Cartesian grid with $N_x=N_y=150$ is employed. 
The acoustic wave equation is solved on the time interval
$[t_0,t_f]=[0,2]$ using $N_t=400$ temporal sampling points. 
The resulting boundary measurements consist of
\[
N_b=2N_x+2(N_y-2)=596
\]
boundary locations. Thus, for each realization, the Dirichlet and Neumann
boundary measurements are represented by
\[
g,h\in\mathbb{R}^{596\times400}.
\]
The initial pressure and squared sound speed are represented by
tensor-product Haar wavelets up to levels $J_f=6,~J_q=5$, resulting in
$N_f=16384,~ N_q=4096$ wavelet coefficients. For each realization, a random
coefficient vector $\beta\in\mathbb{R}^{16384}$ is generated and used to
construct the initial-pressure field. A second independent coefficient
vector is used to generate the fluctuation field $\varepsilon$. The ridge
regularization parameter for the coefficient reconstruction is chosen as
\[
\lambda_\beta=100.
\]
The physical parameters used in the experiments are
\begin{equation}
c_{\min}=1.35,
\qquad
c_{\max}=1.75,
\qquad
c_0=1.54.
\label{eq:physicalparameters}
\end{equation}
This choice allows a relatively wide range of sound-speed variations and
therefore provides a more demanding reconstruction problem than a nearly
homogeneous acoustic medium.

The coupling coefficient is chosen as
\begin{equation}
K
=
0.35\left(c_{\max}^2-c_{\min}^2\right),
\label{eq:Kparameter}
\end{equation}
while the amplitude of the independent acoustic fluctuation is
\begin{equation}
\sigma_\varepsilon
=
0.15\left(c_{\max}^2-c_{\min}^2\right).
\label{eq:sigmaparameter}
\end{equation}
Consequently, the squared sound speed is generated according to
\begin{equation}
q(x)
=
c_0^2
+
K\left(F_{01}(x)-\overline F_{01}\right)
+
\sigma_\varepsilon\varepsilon(x),
\label{eq:q_generation_numerical}
\end{equation}
followed by projection onto the admissible interval
$[c_{\min}^2,c_{\max}^2]$. Here $F_{01}$ denotes the normalized
initial-pressure phantom and $\overline F_{01}$ its spatial mean. The term
\[
K(F_{01}-\overline F_{01})
\]
therefore describes the component of the acoustic heterogeneity that is
correlated with the initial-pressure geometry, whereas
$\sigma_\varepsilon\varepsilon$ introduces an independent acoustic
component. Thus, although the pressure and sound speed are statistically
coupled, the sound-speed field is not determined uniquely by the
initial-pressure field.

The final data set contains $M=5000$ independent realizations. Of these, $4000$ realizations 
are used for training and the remaining $1000$ realizations form an independent test set. 
The test data are not used to fit any of the reconstruction maps. When model parameters are 
selected by validation, a subset of the training data is reserved for this purpose before the 
corresponding regression operator is refitted.

For each pair of boundary measurements $(g,h)$, the frozen convolutional
feature extractor introduced in Section~\ref{sec:cnn} produces a feature
vector $z$. The random convolutional filters and the feature projection are
generated once and remain fixed throughout training and testing. The
coefficient-space reconstruction of Section~\ref{sec:cnn} predicts the
pressure coefficient vector $\widehat{\beta}$ and then uses the
coefficient-space prior $\widehat{\gamma}_0^{\,T}=[1,\widehat{\beta}^{\,T}]T$
to obtain the baseline physical fields
\[
f_0,\qquad q_0,\qquad c_0^{\rm rec}=\sqrt{q_0}.
\]
The performance of the reconstruction framework is evaluated directly in
physical space. For an exact initial pressure $f$ and sound speed $c$, with
corresponding reconstructions $\widehat f$ and $\widehat c$, we define the
relative reconstruction errors
\begin{equation}
E_f
=
\frac{\|f-\widehat f\|_2}{\|f\|_2},
\label{eq:Ef}
\end{equation}
and
\begin{equation}
E_c
=
\frac{\|c-\widehat c\|_2}{\|c\|_2}.
\label{eq:Ec}
\end{equation}

\begin{table}[H]
\centering
\begin{tabular}{llccc}
\hline
Reconstruction & Data set & Mean error & Median error & $R^2$\\
\hline
Pressure untrained CNN+ridge $f_0$
& Training & 0.19521 & 0.19410 & 0.95536\\
& Test     & 0.24616 & 0.24472 & 0.92470\\[1mm]

Sound speed untrained CNN+ridge $c_0^{\rm rec}$
& Training & 0.00828 & 0.00814 & 0.80582\\
& Test     & 0.01013 & 0.00996 & 0.71181\\[1mm]

\hline
\end{tabular}
\caption{Reconstruction performance of the untrained CNN and
ridge-regression framework on the training and held-out test data.
The reported $R^2$ values are computed globally over all samples and
spatial grid points.}
\label{tab:training_test_results}
\end{table}

The results in Table~\ref{tab:training_test_results} establish the
reconstruction performance of the proposed untrained CNN and ridge-regression
framework. On the training set, the initial pressure reconstruction
$f_0$ has a mean/median relative error of $0.19521/0.19410$ and a global
$R^2$ of $0.95536$. The corresponding sound-speed reconstruction
$c_0^{\rm rec}$ has a substantially smaller mean/median relative error of
$0.00828/0.00814$, with a global $R^2$ of $0.80582$. These results show
that the fixed random convolutional features, together with the multiscale
ridge reconstruction and coefficient-space prior, provide accurate
reconstructions of both unknown physical fields without optimization of
the convolutional filters.

The reconstruction accuracy remains high on the held-out test set. For the
initial pressure, the mean/median relative error increases moderately to
$0.24616/0.24472$, while the global $R^2$ remains high at $0.92470$.
For the sound speed, the mean/median relative error is
$0.01013/0.00996$, with a global $R^2$ of $0.71181$. The close agreement
between the mean and median errors for both the training and test sets
indicates that the reported performance is representative of the sample
population rather than being dominated by a small number of realizations.
Although the expected decrease in accuracy from the training to the
held-out test set is more noticeable for the sound speed, the results
demonstrate that the untrained convolutional representation generalizes
to statistically independent test realizations while retaining strong
reconstruction accuracy.

\subsection{Stability with respect to measurement noise}

We next investigate the stability of the already constructed
reconstruction map with respect to perturbations in the boundary
measurements that are not represented during training. Importantly, no
noise augmentation or retraining is performed. The random convolutional
filters, feature projection, normalization parameters, ridge maps, and
coefficient-space prior are constructed entirely from the original
noise-free training set and are subsequently kept fixed. For a prescribed relative noise level $\delta$, independent Gaussian
perturbations are added to the Dirichlet and Neumann measurements
according to
\[
g^\delta
=
g+\delta\frac{\|g\|_F}{\|\xi_g\|_F}\xi_g,
\qquad
h^\delta
=
h+\delta\frac{\|h\|_F}{\|\xi_h\|_F}\xi_h,
\]
where $\xi_g$ and $\xi_h$ have independent standard normal entries.
Consequently,
\[
\frac{\|g^\delta-g\|_F}{\|g\|_F}=\delta,
\qquad
\frac{\|h^\delta-h\|_F}{\|h\|_F}=\delta.
\]
Noise levels of $1\%$, $2\%$, $5\%$, and $10\%$ are considered, with
three independent realizations at each nonzero noise level.
\begin{table}[H]
\centering
\begin{tabular}{ccccc}
\hline
Noise level & Mean $E_f$ & $R_f^2$ & Mean $E_c$ & $R_c^2$\\
\hline
$0\%$  & 0.246162 & 0.924696 & 0.010129 & 0.711812\\
$1\%$  & 0.246163 & 0.924696 & 0.010129 & 0.711806\\
$2\%$  & 0.246167 & 0.924693 & 0.010129 & 0.711814\\
$5\%$  & 0.246203 & 0.924673 & 0.010129 & 0.711788\\
$10\%$ & 0.246337 & 0.924591 & 0.010132 & 0.711642\\
\hline
\end{tabular}
\caption{Stability of the frozen reconstruction map under additive
Gaussian perturbations of the held-out boundary measurements. The nonzero-noise
results are averaged over three independent noise realizations.}
\label{tab:noise_stability}
\end{table}

The results in Table~\ref{tab:noise_stability} demonstrate a high degree
of stability of the frozen reconstruction map. In the absence of noise,
the experiment reproduces the original held-out results, with
$E_f=0.246162$, $R_f^2=0.924696$, $E_c=0.010129$, and
$R_c^2=0.711812$. Even at the largest perturbation level of $10\%$,
these values change only to $E_f=0.246337$, $R_f^2=0.924591$,
$E_c=0.010132$, and $R_c^2=0.711642$, respectively. Thus, a $10\%$
relative perturbation of both boundary data channels produces only a
very small deterioration in reconstruction accuracy. The variation
among the three independent noise realizations is also negligible; at
$10\%$ noise, the standard deviations are approximately
$8\times10^{-6}$ and $1\times10^{-6}$ for the mean pressure and
sound-speed errors, respectively. These observations indicate that the
regularized reconstruction map is empirically stable to substantial
additive measurement perturbations, despite having been constructed
exclusively from noise-free training data.

\subsection{Comparison with a trained CNN feature extractor}
To assess whether training the convolutional filters provides an advantage
over the proposed random-feature construction, we performed an additional
experiment in which the untrained convolutional feature extractor was
replaced by a supervised trained CNN. The remainder of the
coefficient-space reconstruction framework was kept unchanged as far as
possible. In particular, the trained convolutional features were followed
by the same multiscale ridge reconstruction of the pressure coefficients
and the same coefficient-space prior used to reconstruct the squared sound
speed. Thus, the purpose of this experiment is to assess the benefit of
training the convolutional representation rather than to compare with a
different end-to-end reconstruction architecture.

Directly training the CNN to predict the complete pressure coefficient
vector $\beta$ would require a high-dimensional output layer, since the
pressure representation contains $16\,384$ coefficients. We, therefore,
trained the convolutional feature extractor through an auxiliary
physical-space pressure-reconstruction problem. For each training sample,
the exact initial pressure was restricted to a $24\times24$ grid and used
as the supervised target. The resulting trained network has the form
\[
(g,h)
\longrightarrow
{\cal N}_{\rm tr}(g,h)
\longrightarrow
z_{\rm tr}
\longrightarrow
f_{\rm aux},
\]
where ${\cal N}_{\rm tr}$ denotes the trainable convolutional network,
$z_{\rm tr}\in\mathbb{R}^{512}$ is its latent representation, and
$f_{\rm aux}$ is the auxiliary low-resolution pressure prediction. The
network parameters, including the convolutional filters, were optimized
using the Adam algorithm by minimizing the mean-squared error between the
predicted and exact low-resolution pressure fields.

After completion of the supervised training, the auxiliary output layer
was discarded and only the learned latent representation $z_{\rm tr}$ was
retained. The standardized trained features were then supplied to the same
multiscale ridge-regression stage used in the proposed framework,
\[
z_{\rm tr}
\longrightarrow
\widehat{\beta}_{\rm tr}
\longrightarrow
f_{0,\rm tr}.
\]
The corresponding squared sound speed was reconstructed using the same
coefficient-space relation,
\[
\widehat{\gamma}_{\rm tr}^{\,T}
=
[1,\widehat{\beta}_{\rm tr}^{\,T}]T,
\qquad
q_{0,\rm tr}
=
{\cal S}_c(\widehat{\gamma}_{\rm tr}),
\qquad
c_{0,\rm tr}
=
\sqrt{q_{0,\rm tr}}.
\]
Consequently, the principal distinction between the two experiments is
the construction of the convolutional features: the proposed method uses
random convolutional filters generated once and subsequently frozen,
whereas the comparison method learns the convolutional filters from the
training data by backpropagation.

Table~\ref{tab:trained_untrained} compares the held-out reconstruction
accuracy and computational cost of the two approaches. The CPU times
include the complete training and fitting procedure for each method. For
the trained network, this includes CNN optimization, latent-feature
extraction, ridge fitting, and construction of the coefficient-space
prior.

\begin{table}[H]
\centering
\begin{tabular}{llccc}
\hline
Method & Quantity & Train mean/median & Test mean/median & Test $R^2$\\
\hline
Untrained CNN + ridge
& $f_0$ & 0.19521/0.19410 & 0.24616/0.24472 & 0.92470\\
& $c_0$ & 0.00828/0.00814 & 0.01013/0.00996 & 0.71181\\[1mm]

Trained CNN + ridge
& $f_0$ & 0.56853/0.56722 & 0.58519/0.58262 & 0.57635\\
& $c_0$ & 0.01361/0.01351 & 0.01424/0.01380 & 0.25921\\
\hline
\end{tabular}
\caption{Comparison of the untrained and trained convolutional feature
extractors. Relative reconstruction errors are reported as mean/median.
Both approaches use the same multiscale ridge reconstruction and
coefficient-space sound-speed prior.}
\label{tab:trained_untrained}
\end{table}
The comparison shows that supervised training of the convolutional
representation does not improve the reconstruction for the present
architecture. On the training set, the mean/median pressure errors increase
from $0.19521/0.19410$ for the untrained feature extractor to
$0.56853/0.56722$ for the trained CNN, while the corresponding global
$R^2$ decreases from $0.95536$ to $0.60357$. For the sound speed, the
mean/median training errors increase from $0.00828/0.00814$ to
$0.01361/0.01351$, with the global $R^2$ decreasing from $0.80582$ to
$0.47471$. The same behavior is observed on the held-out test set. For the initial
pressure, the mean/median test errors increase from $0.24616/0.24472$ for
the untrained feature extractor to $0.58519/0.58262$ for the trained CNN,
while the corresponding global $R^2$ decreases from $0.92470$ to
$0.57635$. Similarly, for the sound speed, the mean/median test errors
increase from $0.01013/0.00996$ to $0.01424/0.01380$, while the global
$R^2$ decreases from $0.71181$ to $0.25921$. The close agreement between
the mean and median errors within each method indicates that the observed
differences are not driven by a small number of poorly reconstructed
samples. Thus, for the present architecture, learning the convolutional
filters does not produce a latent representation that is more effective
for the subsequent coefficient-space ridge reconstruction. 
One possible explanation is the mismatch between the auxiliary
low-resolution physical-space objective used to train the CNN and the
high-dimensional wavelet-coefficient reconstruction that constitutes
the final inverse task.

The difference in computational cost is also substantial. The proposed
untrained framework requires approximately $1.26$ hrs of CPU time, whereas
the trained-CNN framework requires approximately $7.52$ hrs. The latter is
therefore approximately $5.97$ times more expensive computationally. In
the trained case, $415.28$ min are required for CNN optimization alone,
with an additional $23.45$ min for feature extraction and $12.77$ min for
ridge regression and construction of the coefficient-space prior. Hence,
the use of fixed random convolutional features eliminates the dominant
iterative network-training stage while simultaneously providing better
reconstruction accuracy in the present experiments. These results provide 
computational motivation for the use of an untrained convolutional feature 
extractor in the proposed framework. The combination of fixed random 
convolutional features and ridge regression therefore provides a particularly efficient 
alternative for the joint reconstruction problem considered in this work.

\subsection{Reconstruction of unseen phantom geometries}
\label{subsec:unseen_phantoms}

We next examine the performance of the reconstruction framework on
phantom geometries that are not represented in the training data. Three
deterministic examples are considered: a disk phantom, a smooth Gaussian
phantom, and a heart--lung phantom. These cases provide spatial structures
that differ from the random phantom realizations used to construct the
training and held-out test sets.

For each unseen phantom, synthetic boundary measurements are generated by
solving the same acoustic forward problem on $\Omega=[-1,1]^2$
using the same $150\times150$ spatial discretization, $400$ temporal
sampling points, and $596$ boundary measurement locations employed in the
training and test experiments. The admissible sound-speed range and
background sound speed are also unchanged,
\[
c_{\min}=1.35,
\qquad
c_{\max}=1.75,
\qquad
c_0=1.54,
\]
and the same coupling coefficient
\[
K
=
0.35(c_{\max}^2-c_{\min}^2)
\]
is employed.

For these experiments, the independent acoustic perturbation is suppressed
in order to isolate the effect of changing the phantom morphology. The
squared sound speed is therefore generated according to
\begin{equation}
q(x)
=
{\cal P}_{[c_{\min}^2,c_{\max}^2]}
\left[
c_0^2
+
K\left(F_{01}(x)-\overline F_{01}\right)
\right].
\label{eq:q_unseen}
\end{equation}

The first example is a disk phantom consisting of a single
piecewise-constant circular inclusion of radius $0.34$, centered at
$(0.05,-0.08)$. After normalization, the inclusion has uniform initial
pressure within the disk and zero pressure outside. This simple,
sharply bounded geometry provides a controlled test of the ability of the
reconstruction maps to recover an isolated structure that was not
explicitly represented in the training data. In addition to the clean
measurements, we test the frozen untrained CNN+ridge reconstruction after
adding $10\%$ relative Gaussian noise independently to the Dirichlet and
Neumann boundary measurements.

\begin{figure}[H]
\centering

% Initial pressure
\subfloat[True $f$]{
\includegraphics[width=0.235\textwidth]{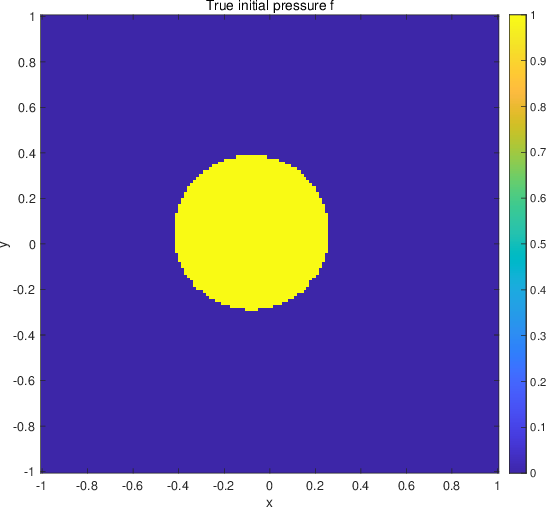}}
\hfill
\subfloat[Untrained CNN+ridge, clean]{
\includegraphics[width=0.235\textwidth]{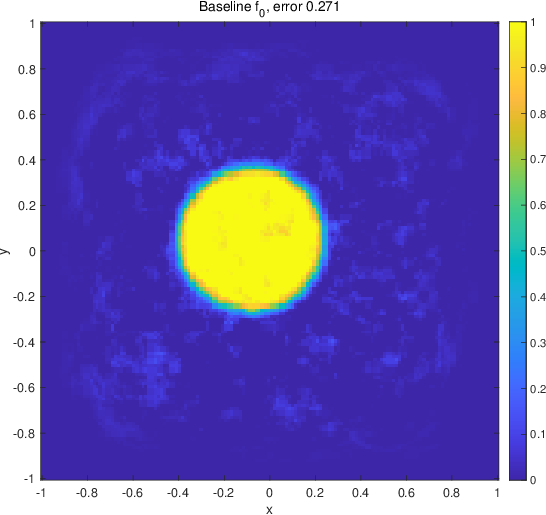}}
\hfill
\subfloat[Untrained CNN+ridge, $10\%$ noise]{
\includegraphics[width=0.235\textwidth]{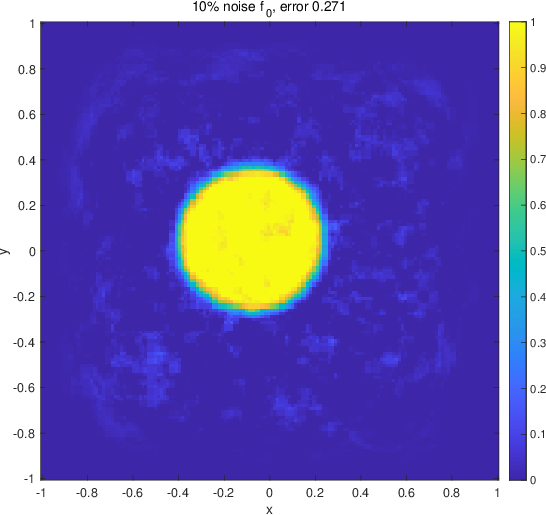}}
\hfill
\subfloat[Trained CNN+ridge, clean]{
\includegraphics[width=0.235\textwidth]{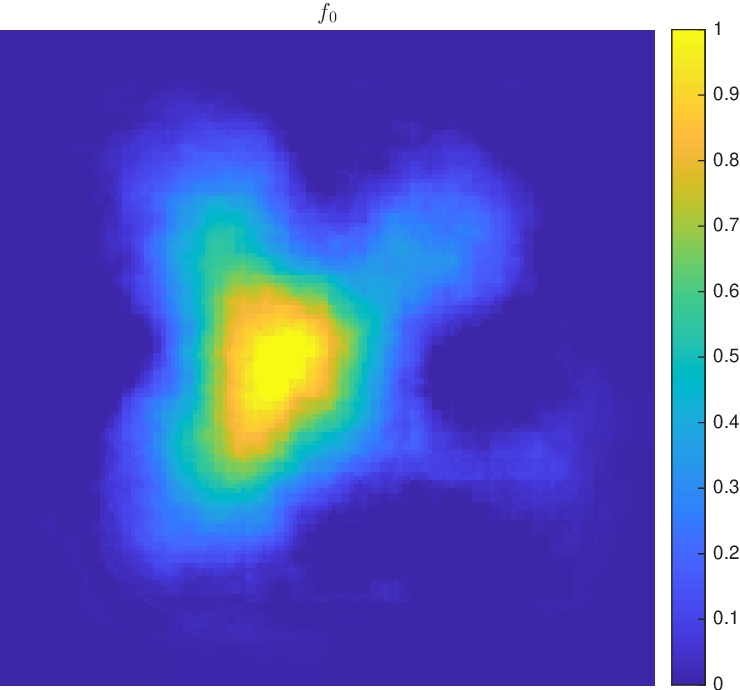}}\\

% Sound speed
\subfloat[True $c$]{
\includegraphics[width=0.235\textwidth]{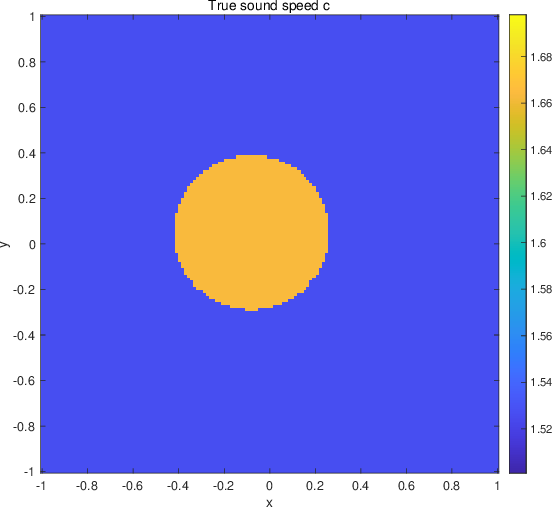}}
\hfill
\subfloat[Untrained CNN+ridge, clean]{
\includegraphics[width=0.235\textwidth]{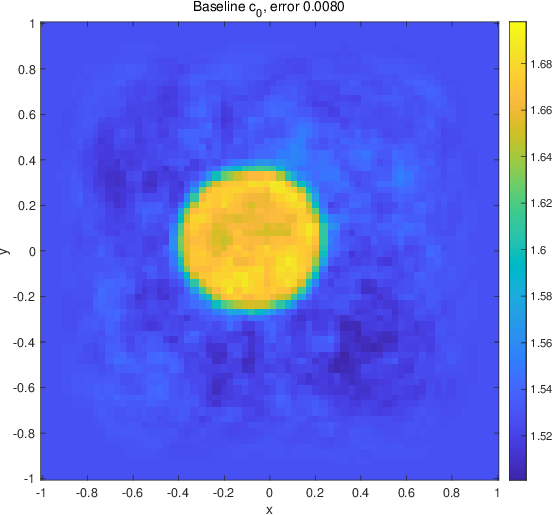}}
\hfill
\subfloat[Untrained CNN+ridge, $10\%$ noise]{
\includegraphics[width=0.235\textwidth]{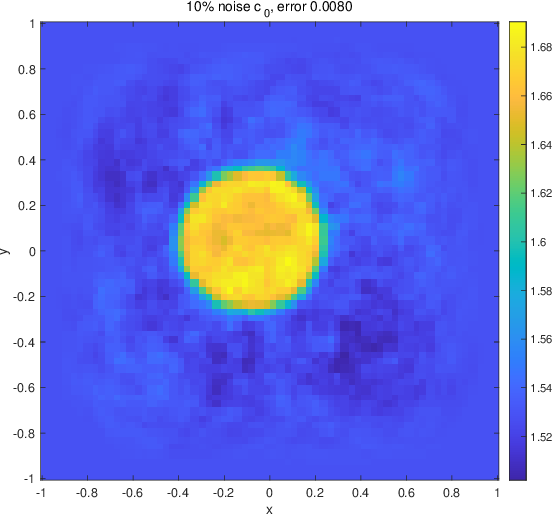}}
\hfill
\subfloat[Trained CNN+ridge, clean]{
\includegraphics[width=0.235\textwidth]{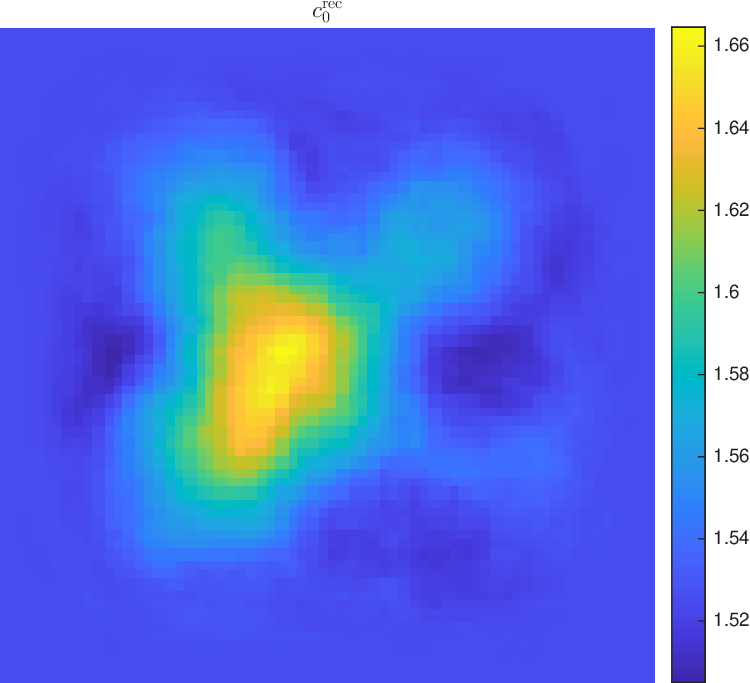}}

\caption{Reconstruction of the unseen disk phantom. The first column
shows the true initial pressure and sound speed. The second and third
columns show the reconstructions obtained using the proposed untrained
CNN+ridge method from clean measurements and measurements contaminated
by $10\%$ relative Gaussian noise, respectively. The fourth column shows
the reconstruction obtained using the trained CNN+ridge method from
clean measurements.}
\label{fig:unseen_disk}
\end{figure}

The second example is a Gaussian phantom centered at $(-0.08,0.10)$.
The initial pressure is defined by an anisotropic Gaussian profile with
characteristic widths $\sigma_x=0.23$ and $\sigma_y=0.16$, rotated through
an angle $\pi/6$. In contrast with the sharply bounded disk and the
piecewise-defined structures used in the training ensemble, this phantom
has a smooth, continuously varying spatial profile. It therefore tests
whether the reconstruction framework can generalize from localized
piecewise structures to a smooth and spatially anisotropic pressure
distribution.

\begin{figure}[H]
\centering

% Initial pressure
\subfloat[True $f$]{
\includegraphics[width=0.235\textwidth]{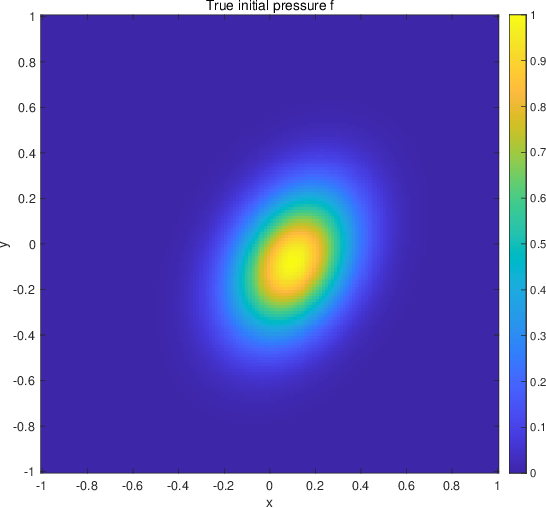}}
\hfill
\subfloat[Untrained CNN+ridge, clean]{
\includegraphics[width=0.235\textwidth]{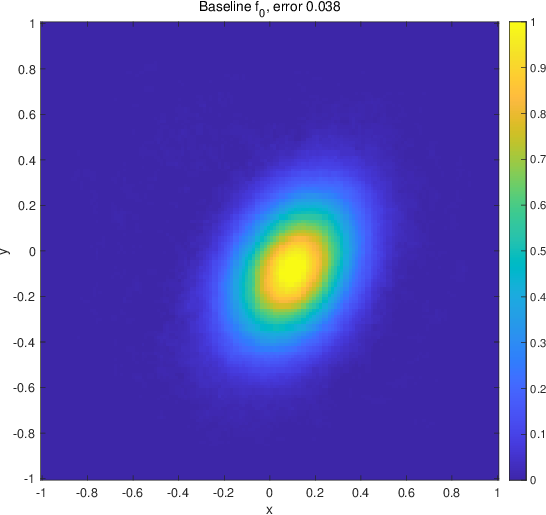}}
\hfill
\subfloat[Untrained CNN+ridge, $10\%$ noise]{
\includegraphics[width=0.235\textwidth]{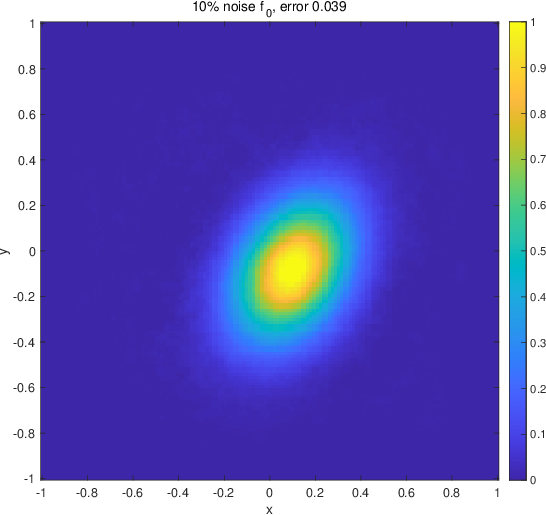}}
\hfill
\subfloat[Trained CNN+ridge, clean]{
\includegraphics[width=0.235\textwidth]{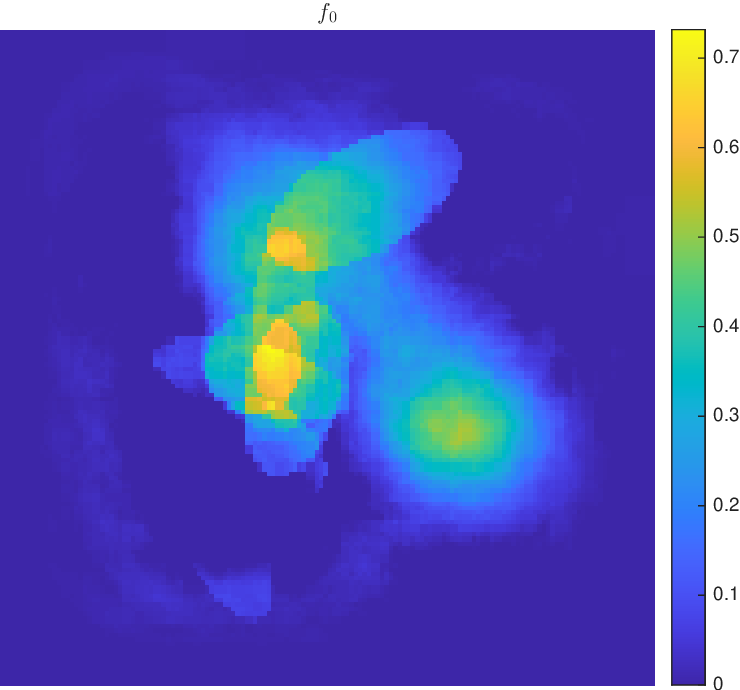}}\\

% Sound speed
\subfloat[True $c$]{
\includegraphics[width=0.235\textwidth]{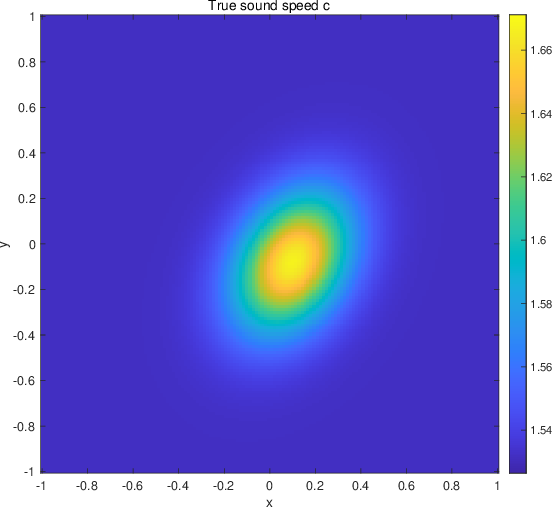}}
\hfill
\subfloat[Untrained CNN+ridge, clean]{
\includegraphics[width=0.235\textwidth]{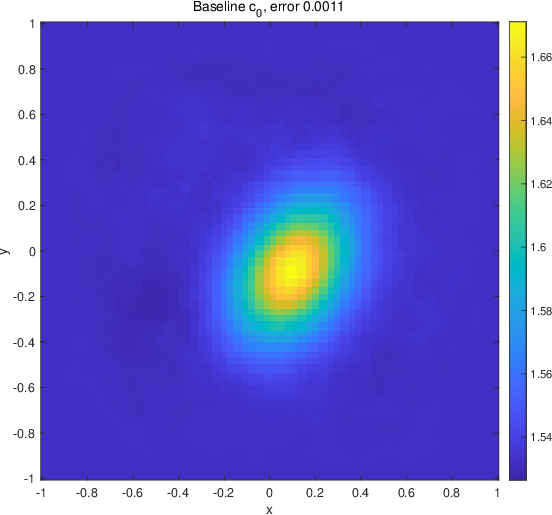}}
\hfill
\subfloat[Untrained CNN+ridge, $10\%$ noise]{
\includegraphics[width=0.235\textwidth]{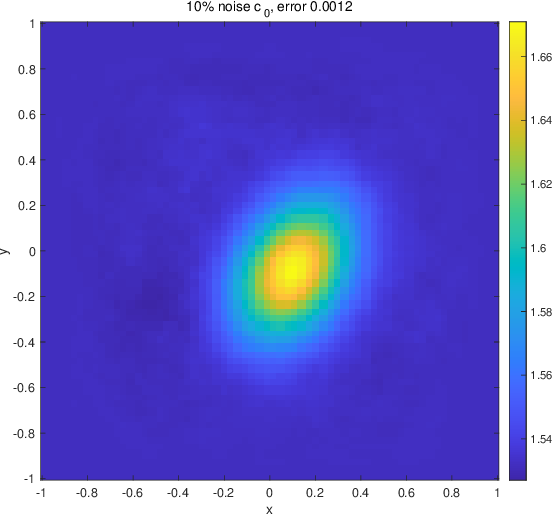}}
\hfill
\subfloat[Trained CNN+ridge, clean]{
\includegraphics[width=0.235\textwidth]{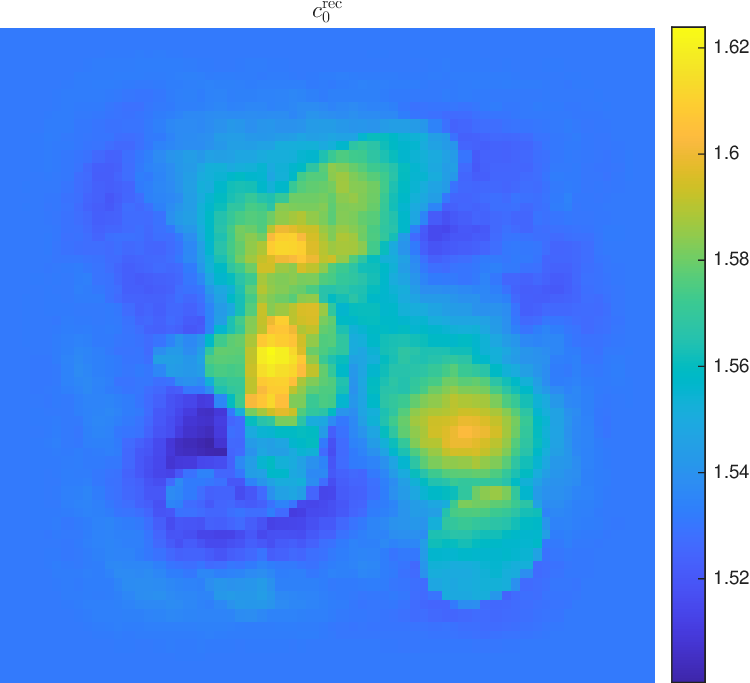}}

\caption{Reconstruction of the unseen Gaussian phantom. The first column
shows the true initial pressure and sound speed. The second and third
columns show the reconstructions obtained using the proposed untrained
CNN+ridge method from clean measurements and measurements contaminated
by $10\%$ relative Gaussian noise, respectively. The fourth column shows
the reconstruction obtained using the trained CNN+ridge method from
clean measurements.}
\label{fig:unseen_gaussian}
\end{figure}

The third example is a simplified heart--lung phantom consisting of three
elliptical structures. Two larger ellipses, centered approximately at
$(0.5,0)$ and $(-0.6,0)$ and rotated by angles $-\pi/7$ and $\pi/7$,
respectively, represent the two lungs. A smaller ellipse centered near
$(-0.1,0.4)$ represents the heart and is assigned one-half the amplitude
of the lung regions. This phantom introduces a structured multi-object
geometry containing objects of different sizes, orientations, and
amplitudes, and hence provides a more complex test of generalization than
either the single disk or the smooth Gaussian phantom.

\begin{figure}[H]
\centering

% Initial pressure
\subfloat[True $f$]{
\includegraphics[width=0.235\textwidth]{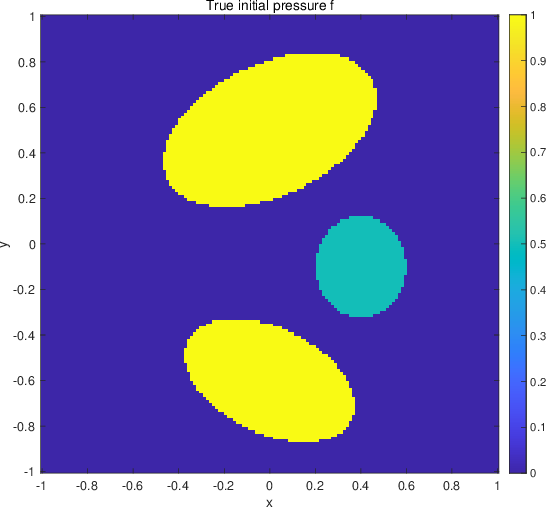}}
\hfill
\subfloat[Untrained CNN+ridge, clean]{
\includegraphics[width=0.235\textwidth]{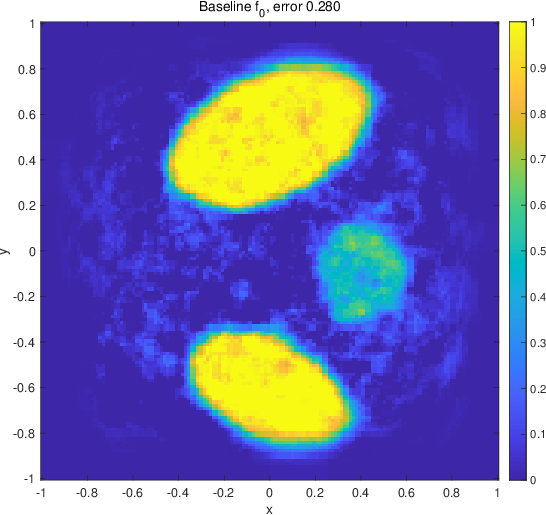}}
\hfill
\subfloat[Untrained CNN+ridge, $10\%$ noise]{
\includegraphics[width=0.235\textwidth]{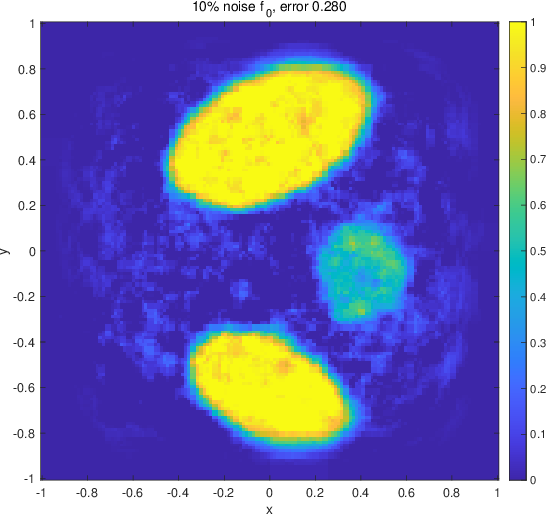}}
\hfill
\subfloat[Trained CNN+ridge, clean]{
\includegraphics[width=0.235\textwidth]{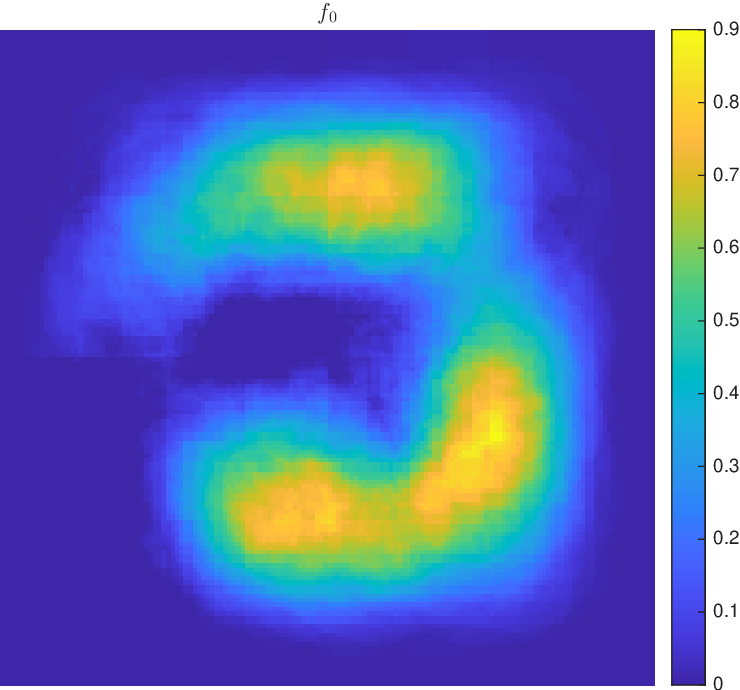}}\\

% Sound speed
\subfloat[True $c$]{
\includegraphics[width=0.235\textwidth]{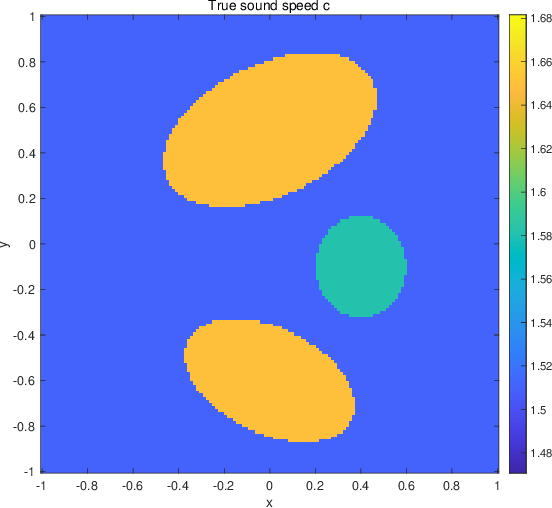}}
\hfill
\subfloat[Untrained CNN+ridge, clean]{
\includegraphics[width=0.235\textwidth]{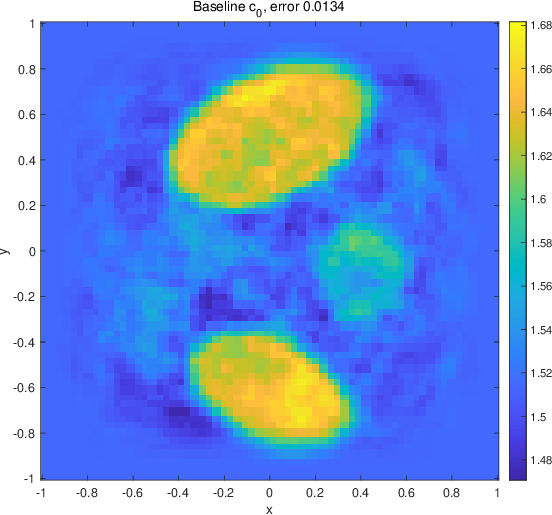}}
\hfill
\subfloat[Untrained CNN+ridge, $10\%$ noise]{
\includegraphics[width=0.235\textwidth]{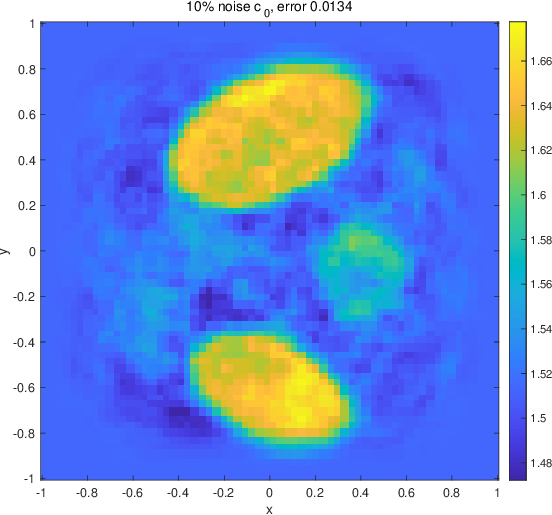}}
\hfill
\subfloat[Trained CNN+ridge, clean]{
\includegraphics[width=0.235\textwidth]{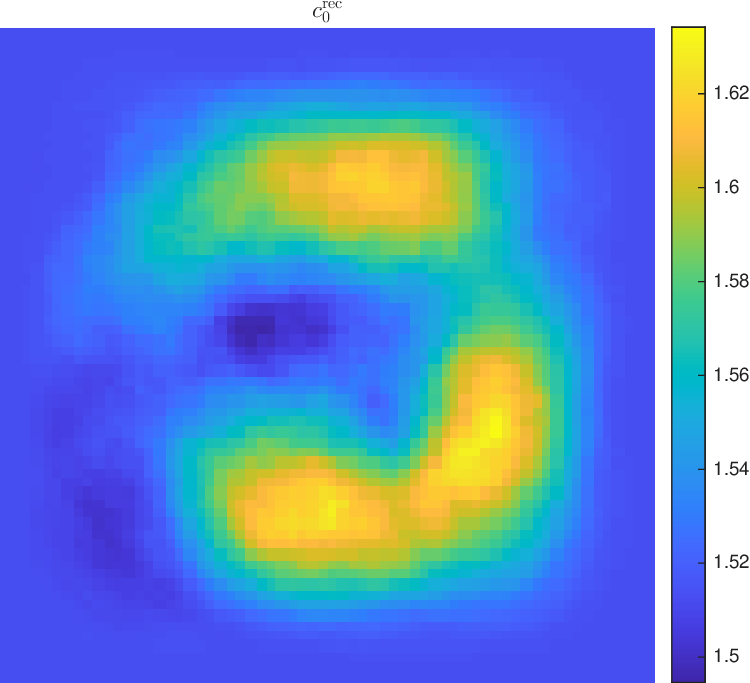}}

\caption{Reconstruction of the unseen heart--lung phantom. The first column
shows the true initial pressure and sound speed. The second and third
columns show the reconstructions obtained using the proposed untrained
CNN+ridge method from clean measurements and measurements contaminated
by $10\%$ relative Gaussian noise, respectively. The fourth column shows
the reconstruction obtained using the trained CNN+ridge method from
clean measurements.}
\label{fig:unseen_heart_lung}
\end{figure}

The corresponding relative errors are calculated using
\eqref{eq:Ef}--\eqref{eq:Ec}. For the noisy experiment, independent
Gaussian perturbations are added to the two boundary measurements
according to
\[
g^\delta
=
g+\delta\frac{\|g\|_F}{\|\xi_g\|_F}\xi_g,
\qquad
h^\delta
=
h+\delta\frac{\|h\|_F}{\|\xi_h\|_F}\xi_h,
\qquad
\delta=0.10,
\]
where $\xi_g$ and $\xi_h$ have independent standard normal entries.
The results are given in Table \ref{tab:phantom_results}.

\begin{table}[H]
\centering
\begin{tabular}{llcccc}
\hline
Phantom & Method/data & $E_f$ & $R_f^2$ & $E_c$ & $R_c^2$\\
\hline

Disk
& Untrained, clean
& 0.270509 & 0.919631 & 0.007996 & 0.899434\\
& Untrained, $10\%$ noise
& 0.270908 & 0.919393 & 0.008011 & 0.899053\\
& Trained, clean
& 0.623817 & 0.572596 & 0.017175 & 0.536073\\[1mm]

Gaussian
& Untrained, clean
& 0.038391 & 0.998336 & 0.001138 & 0.993573\\
& Untrained, $10\%$ noise
& 0.039248 & 0.998261 & 0.001152 & 0.993409\\
& Trained, clean
& 0.945115 & -0.008256 & 0.014777 & -0.084208\\[1mm]

Heart--lung
& Untrained, clean
& 0.280243 & 0.900608 & 0.013369 & 0.852201\\
& Untrained, $10\%$ noise
& 0.280198 & 0.900640 & 0.013384 & 0.851877\\
& Trained, clean
& 0.697759 & 0.383842 & 0.027277 & 0.384720\\
\hline
\end{tabular}
\caption{Reconstruction performance on the three completely unseen
phantom geometries. The untrained CNN+ridge method is evaluated using
both clean measurements and measurements contaminated by $10\%$
relative Gaussian noise. The noisy results are averages over three
independent realizations. For comparison, the clean-measurement results
obtained using the trained CNN+ridge method are also reported.}
\label{tab:phantom_results}
\end{table}

For the disk phantom, the untrained CNN+ridge reconstruction successfully
recovers the location, size, and sharply defined circular geometry of the
inclusion, with only moderate artifacts in the surrounding background.
The corresponding sound-speed reconstruction also preserves the circular
structure and its contrast relative to the background. This behavior is
essentially unchanged when $10\%$ noise is added to the measurements:
$E_f$ changes only from $0.270509$ to $0.270908$, while $R_f^2$ changes
from $0.919631$ to $0.919393$. Similarly, $E_c$ changes from $0.007996$
to $0.008011$ and $R_c^2$ from $0.899434$ to $0.899053$. In comparison,
the trained CNN reconstruction from clean measurements is substantially
less accurate and produces broad, diffuse structures rather than the
single localized disk.

The difference between the two convolutional approaches is particularly
evident for the Gaussian phantom. The untrained CNN+ridge method
reconstructs the smooth anisotropic profile with very high fidelity,
preserving its location, orientation, and continuously varying amplitude.
The reconstruction also remains highly accurate after substantial
measurement perturbation. At $10\%$ noise, the pressure error is
$0.039248$, compared with $0.038391$ for clean measurements, while
$R_f^2$ remains $0.998261$. The sound-speed reconstruction changes from
$E_c=0.001138$ and $R_c^2=0.993573$ to $E_c=0.001152$ and
$R_c^2=0.993409$. In contrast, the trained CNN produces several separated
and overlapping structures rather than the single smooth Gaussian profile,
with correspondingly poor quantitative accuracy.

For the more complicated heart--lung phantom, the untrained CNN+ridge
method again preserves the principal geometry of the unseen object. The
two larger elliptical regions and the smaller central region remain
identifiable in both the pressure and sound-speed reconstructions,
although some background artifacts and smoothing are present. The
addition of $10\%$ measurement noise has negligible effect: the pressure
error changes from $0.280243$ to $0.280198$ and the sound-speed error from
$0.013369$ to $0.013384$, while the corresponding $R^2$ values remain
essentially unchanged. The very small favorable change in the pressure
metrics should not be interpreted as an improvement due to noise, but
rather as variation associated with the particular noise realizations.
The trained CNN, by comparison, recovers only a coarse multi-lobed
structure and substantially blurs the individual objects.

Taken together, these experiments demonstrate two distinct aspects of
generalization. First, the fixed random features used by the untrained
CNN+ridge framework retain sufficient spatial information for the
coefficient-space reconstruction to recover geometries ranging from a
sharply bounded disk to a smooth Gaussian profile and a structured
multi-object phantom. Second, this out-of-distribution performance is
empirically stable under substantial measurement perturbations that were
not represented during construction of the reconstruction map. Even at
$10\%$ relative noise, the reconstruction metrics for all three unseen
phantoms remain close to their clean-measurement values. In contrast, the
supervised CNN features obtained with the present training procedure do
not transfer effectively to these unseen geometries even for clean data.
A likely explanation is that the trained CNN is optimized for the
auxiliary low-resolution pressure-reconstruction objective rather than
directly for recovery of the high-dimensional wavelet coefficient vector
used by the subsequent ridge stage. Its latent representation is therefore
more specialized to the training objective, whereas the fixed random
convolutional features provide a more transferable representation for the
coefficient-space reconstruction. Since the sound-speed coefficients are
subsequently inferred from the reconstructed pressure coefficients through
the learned coefficient-space prior, deterioration of the trained-CNN
pressure reconstruction is also propagated to the reconstructed
sound-speed field.

\section{Conclusion}

We have presented a regularized framework for the joint reconstruction
of the initial pressure and spatially varying sound speed in
photoacoustic tomography. The method combines finite-dimensional
wavelet representations, fixed nonlinear features generated by an
untrained CNN, Tikhonov-regularized recovery of the pressure
coefficients, and a coefficient-space prior relating the initial
pressure to the squared sound speed. Numerical experiments demonstrate
accurate reconstruction on held-out data as well as strong
generalization to previously unseen phantom geometries. The
reconstruction map also remains stable when the boundary measurements
are contaminated by noise not represented during training, with only
minor changes in reconstruction accuracy for noise levels up to $10\%$
on both the held-out data and the unseen phantoms. For the supervised
trained-CNN comparator considered here, training the convolutional
representation did not improve reconstruction accuracy and required
approximately six times the computational cost of the untrained
approach. These results indicate that fixed nonlinear measurement
features combined with coefficient-space regularization provide an
effective, computationally efficient, and empirically stable strategy
for the joint photoacoustic inverse problem. Future work will
investigate robustness to model mismatch, more realistic acoustic
models, and experimental and three-dimensional photoacoustic data.

\section*{Acknowledgements}
S. Roy was partly supported by the US National Science Foundation grant no. 2309491.

\bibliographystyle{unsrt}
\bibliography{literatureJointPAT}

@article{lan2021compressed,
  author  = {Lan, Hengrong and Zhang, Juze and Yang, Changchun and Gao, Fei},
  title   = {Compressed sensing for photoacoustic computed tomography
             based on an untrained neural network with a shape prior},
  journal = {Biomedical Optics Express},
  volume  = {12},
  number  = {12},
  pages   = {7835--7848},
  year    = {2021},
  doi     = {10.1364/BOE.441901}
}

@article{suhonen2025multiple,
  author  = {Suhonen, Miika and Lucka, Felix and Pulkkinen, Aki and
             Arridge, Simon},
  title   = {Reconstructing multiple initial pressure and speed of sound
             distributions simultaneously in photoacoustic tomography},
  journal = {Photoacoustics},
  volume  = {45},
  pages   = {100748},
  year    = {2025},
  doi     = {10.1016/j.pacs.2025.100748}
}

@article{cai2019feature,
  author  = {Cai, Chuangjian and Wang, Xuanhao and Si, Ke and
             Qian, Jun and Luo, Jianwen and Ma, Cheng},
  title   = {Feature coupling photoacoustic computed tomography for joint
             reconstruction of initial pressure and sound speed in vivo},
  journal = {Biomedical Optics Express},
  volume  = {10},
  number  = {7},
  pages   = {3447--3462},
  year    = {2019},
  doi     = {10.1364/BOE.10.003447}
}

@article{shan2019simultaneous,
  author  = {Shan, Hongming and Wiedeman, Christopher and
             Wang, Ge and Yang, Yang},
  title   = {Simultaneous reconstruction of the initial pressure and
             sound speed in photoacoustic tomography using a
             deep-learning approach},
  journal = {arXiv preprint arXiv:1907.09951},
  year    = {2019}
}

@article{arridge2016adjoint,
  author  = {Arridge, Simon R. and Betcke, Marta M. and Cox, Ben T. and
             Lucka, Felix and Treeby, Bradley E.},
  title   = {On the adjoint operator in photoacoustic tomography},
  journal = {Inverse Problems},
  volume  = {32},
  number  = {11},
  pages   = {115012},
  year    = {2016}
}

@article{matthews2018parameterized,
  author  = {Matthews, Thomas P. and Poudel, Joemini and Li, Lei and
             Wang, Lihong V. and Anastasio, Mark A.},
  title   = {Parameterized joint reconstruction of the initial pressure
             and sound speed distributions for photoacoustic computed
             tomography},
  journal = {SIAM Journal on Imaging Sciences},
  volume  = {11},
  number  = {2},
  pages   = {1560--1588},
  year    = {2018},
  doi     = {10.1137/17M1153649}
}

@article{dey2024high,
  title={A high contrast and resolution reconstruction algorithm in quantitative photoacoustic tomography},
  author={Dey, Anwesa and Borz{\`\i}, Alfio and Roy, Souvik},
  journal={Journal of Computational and Applied Mathematics},
  volume={451},
  pages={116065},
  year={2024},
  publisher={Elsevier}
}

@article{roy2025pinn,
  title={A PINN-driven game-theoretic framework in limited data photoacoustic tomography},
  author={Roy, Souvik and Pal, Suvra},
  journal={Inverse Problems},
  volume={41},
  number={11},
  pages={115011},
  year={2025},
  publisher={IOP Publishing}
}

@article{matthews2017joint,
  author  = {Matthews, Thomas P. and Anastasio, Mark A.},
  title   = {Joint reconstruction of the initial pressure and speed of
             sound distributions from combined photoacoustic and
             ultrasound tomography measurements},
  journal = {Inverse Problems},
  volume  = {33},
  number  = {12},
  pages   = {124002},
  year    = {2017},
  doi     = {10.1088/1361-6420/aa9384}
}

@article{hauptmann2020deep,
  title={Deep learning in photoacoustic tomography: Current approaches
         and future directions},
  author={Hauptmann, Andreas and Cox, Ben},
  journal={Journal of Biomedical Optics},
  volume={25},
  number={11},
  pages={112903},
  year={2020},
  doi={10.1117/1.JBO.25.11.112903}
}

@article{grohl2021deep,
  title={Deep learning for biomedical photoacoustic imaging: A review},
  author={Gr{\"o}hl, Janek and Schellenberg, Melanie and Dreher, Kris and
          Maier-Hein, Lena},
  journal={Photoacoustics},
  volume={22},
  pages={100241},
  year={2021},
  doi={10.1016/j.pacs.2021.100241}
}

@article{dispirito2021sounding,
  title={Sounding out the hidden data: A concise review of deep learning
         in photoacoustic imaging},
  author={DiSpirito, Anthony, III and Vu, Tri and Pramanik, Manojit and
          Yao, Junjie},
  journal={Experimental Biology and Medicine},
  volume={246},
  number={12},
  pages={1355--1367},
  year={2021},
  doi={10.1177/15353702211000310}
}

@article{xu2006photoacoustic,
  author  = {Xu, Minghua and Wang, Lihong V.},
  title   = {Photoacoustic imaging in biomedicine},
  journal = {Review of Scientific Instruments},
  volume  = {77},
  number  = {4},
  pages   = {041101},
  year    = {2006}
}

@article{beard2011biomedical,
  author  = {Beard, Paul},
  title   = {Biomedical photoacoustic imaging},
  journal = {Interface Focus},
  volume  = {1},
  number  = {4},
  pages   = {602--631},
  year    = {2011}
}

@inproceedings{cox2009photoacoustic,
  author    = {Cox, B. T. and Laufer, J. G. and Beard, P. C.},
  title     = {The challenges for quantitative photoacoustic imaging},
  booktitle = {Proceedings of SPIE},
  volume    = {7177},
  year      = {2009}
}

@article{stefanov2009thermoacoustic,
  author  = {Stefanov, Plamen and Uhlmann, Gunther},
  title   = {Thermoacoustic tomography with variable sound speed},
  journal = {Inverse Problems},
  volume  = {25},
  number  = {7},
  pages   = {075011},
  year    = {2009}
}

@article{stefanov2013instability,
  author  = {Stefanov, Plamen and Uhlmann, Gunther},
  title   = {Instability of the linearized problem in multiwave tomography
             of recovery both the source and the speed},
  journal = {Inverse Problems and Imaging},
  volume  = {7},
  number  = {4},
  pages   = {1367--1377},
  year    = {2013},
  doi     = {10.3934/ipi.2013.7.1367}
}

@article{huang2013full,
  author  = {Huang, Chao and Wang, Kun and Nie, Liming and Wang, Lihong V. and Anastasio, Mark A.},
  title   = {Full-wave iterative image reconstruction in photoacoustic tomography with acoustically inhomogeneous media},
  journal = {IEEE Transactions on Medical Imaging},
  volume  = {32},
  number  = {6},
  pages   = {1097--1110},
  year    = {2013}
}

@book{daubechies1992ten,
  author    = {Daubechies, Ingrid},
  title     = {Ten Lectures on Wavelets},
  publisher = {SIAM},
  address   = {Philadelphia},
  year      = {1992}
}

@book{mallat2009wavelet,
  author    = {Mallat, St{\'e}phane},
  title     = {A Wavelet Tour of Signal Processing},
  edition   = {3},
  publisher = {Academic Press},
  address   = {Amsterdam},
  year      = {2009}
}

@inproceedings{saxe2011random,
  author    = {Saxe, Andrew and Koh, Pang Wei and Chen, Zhenghao and Bhand, Maneesh and Suresh, Bipin and Ng, Andrew},
  title     = {On random weights and unsupervised feature learning},
  booktitle = {Proceedings of the 28th International Conference on Machine Learning (ICML)},
  pages     = {1089--1096},
  year      = {2011}
}

@inproceedings{rahimi2008random,
  author    = {Rahimi, Ali and Recht, Benjamin},
  title     = {Random features for large-scale kernel machines},
  booktitle = {Advances in Neural Information Processing Systems},
  volume    = {21},
  pages     = {1177--1184},
  year      = {2008}
}

\end{document}